%% file: main.tex
\documentclass{article} 

\usepackage{url}

\usepackage[english]{babel}

\usepackage[letterpaper,top=2cm,bottom=2cm,left=3cm,right=3cm,marginparwidth=1.75cm]{geometry}

\usepackage{amsmath}
\usepackage{graphicx}
\usepackage[colorlinks=true, allcolors=blue]{hyperref}

\title{ Artificial Intelligence: $70$ Years Down the Road }

 \date{}

\author{Lin Zhang \\
International Digital Economy Academy (IDEA)\\
\texttt{zhanglin@idea.edu.cn}
}

\begin{document}
\maketitle

\begin{abstract}
Artificial intelligence (AI) has a history of nearly a century from its inception to the present day. We have summarized the development trends and discovered universal rules, including both success and failure. We have analyzed the reasons from both technical and philosophical perspectives to help understand the reasons behind the past failures and current successes of AI, and to provide a basis for thinking and exploring future development. Specifically, we have found that the development of AI in different fields, including computer vision, natural language processing, and machine learning, follows a pattern from rules to statistics to data-driven methods. In the face of past failures and current successes, we need to think systematically about the reasons behind them. Given the unity of AI between natural and social sciences, it is necessary to incorporate philosophical thinking to understand and solve AI problems, and we believe that starting from the dialectical method of Marx is a feasible path. We have concluded that the sustainable development direction of AI should be human-machine collaboration and a technology path centered on computing power. Finally, we have summarized the impact of AI on society from this trend.

\end{abstract}

\input{tex/1_intro.tex}
\input{tex/2_AI_developmement.tex}

\input{tex/4_Phi}

\input{tex/5_social}

\input{tex/conclusion.tex}

\bibliographystyle{alpha}
\bibliography{sample}

\end{document}

%% file: tex/1_intro.tex
\section{Introduction}

Since the concept of AI was introduced in the 1950s~\cite{McCarthyMRS06}, researchers have been striving to make machines intelligent, leading to important sub-fields such as computer vision, natural language processing, and machine learning. Despite ups and downs, AI technology has been rapidly advancing, especially in the last two decades with the success of deep learning. However, the pursuit of a comprehensive understanding of AI mechanisms and how to accelerate its evolution continues. With more data, computing power, and minds than ever before, it is time to review the development of AI, as the reasons behind the ups and downs of its history are worth exploring. Otherwise, AI is likely to repeat the tragedies of the past.

Here, we reflect on the development of AI over the years and how past developments may extend into the future. We first analyze the technological changes in the core areas of AI over the past century. Although the research content and methods of sub-fields differ greatly, their development logic is remarkably similar, and can be summarized as the path from rules to statistics and then to data-driven approaches. The dilemma of AI development, from a technical perspective, is fundamentally due to the lack of mathematical tools. To further explore the reasons for this development model, we analyze it from a philosophical perspective. Specifically, we analyze the problems of AI from the perspectives of dialectics and pragmatism, and believe that AI needs to be understood and solved from the perspectives of contradiction and practice based on Marxist dialectics. Based on this, we believe that human-machine interaction is the inevitable path of AI development, and that AI technology should revolve around the development of computing power. Finally, we outline the changes that AI brings to society.

%% file: tex/2_AI_developmement.tex
\section{A Brief History of AI Technology}

AI research covers a wide range of fields, and we have chosen the most representative research areas, including computer vision, natural language processing, and machine learning, to understand their development history. We will find that from a technical perspective, the dilemma of AI is fundamentally due to the lack of mathematical ability, and we are still unable to effectively model the complex real world. Ultimately, we rely on empirical methods to compensate for the shortcomings of mathematics.

\subsection{Computer Vision}
Computer vision has a history of more than 60 years since its establishment in the 1960s~\cite{CV50year}, and it is a major branch of artificial intelligence. Its core task is to understand the content of images. From the 1960s to the 2000s, the development of computer vision can basically be attributed to a core idea: structured combination. That is to say, past researchers believed that the objects in images were composed of some basic structural units combined in a certain way, and representative figures include David Marr~\cite{Marr1982} and King-Sun Fu. Specifically, Marr started from neuroscience and believed that the process of visual recognition was completed through three progressive levels of visual features, namely 2D (such as node, edge information, etc.), 2.5D (the aggregation of 2D information into 3D contour information) and 3D information. Marr's computer vision theory imitated the process of human vision. He believed that the calculation methods of the brain and the computer were the same. The three-stage features he proposed were a simulation of human vision, which had a dominant influence on the development of computer vision for many years. Researchers have proposed a series of methods for different stage features. For 2D information, representative works include the Scale-invariant Feature Transform and the Canny edge detector for edge extraction, to support downstream visual tasks such as classification, recognition, and segmentation. The analysis of 2.5D and 3D visual information has become a key research topic in the early development of computer vision. For example, the first specialized computer vision paper (from Larry Roberts) analyzed the 3D structural information of objects in images, and similar ideas appeared in the cylinder structure research paradigm proposed by Thomas Binford~\cite{Binford1992} later on. However, the limitations of this type of research were too great, and the complexity of real visual objects made it difficult to complete them by simple geometric combinations. With the success of SIFT feature extraction, this type of research gradually withdrew from the historical stage.

Fu believed that there was a higher-level grammar structure in the world, so understanding images could be achieved by learning this grammar structure. Compared to Marr's three clear levels, Fu's method was more general, and therefore more difficult to design effective computational methods. Therefore, there were few followers until the beginning of this century, when Song-Chun Zhu~\cite{Zhu2006} made some progress in image segmentation using this kind of thinking. However, subsequent research was still rare, mainly because it was difficult to model such grammar structures.

The 2000s were a transitional period for computer vision. The mainstream method was based on statistical models of hand-crafted features, and a series of highly practical works were produced during this period, such as the Viola-Jones algorithm~\cite{ViolaJones} for face detection, the histograms of oriented gradient (HOG) descriptors~\cite{Triggs_hog} and its improved Deformable Parts Model (DPM) algorithm~\cite{felzenszwalb2009object} for pedestrian detection, and the Spatial Pyramid Matching algorithm~\cite{Lazebnik06} for image matching and recognition. It should be emphasized that the success of these algorithms was based on the expert-designed features for image understanding, generally utilizing the properties of gradient distribution on the image. These studies gradually moved away from Marr's three-stage cognition. From a computable perspective, feature-based visual research brought about significant performance improvements, which laid the foundation for subsequent deep learning-based research, because the latter is a data-driven feature rather than a hand-designed one.

The concept of deep learning~\cite{Hinton2006} was proposed by Geoffrey Hinton and his students in 2006, but it was not until 2012 that AlexNet's~\cite{Alex2012} outstanding performance on the ImageNet dataset attracted public attention. Deep learning algorithms, represented by convolutional neural networks (CNNs), brought computer vision into the era of deep learning, a new paradigm driven by data and computing power. With the increase in model parameters, deep learning has brought significant improvements in traditional computer vision tasks. Applications such as face recognition have reached commercial-grade capabilities. Finally, computer vision research converged on image representation learning, from traditional key point localization and representation to block-based representation learning represented by Vision Transformer (ViT) ~\cite{ViT2020}, which completely overturned the visual theory since Marr.

At this point, we may need to reflect on the ideological foundation of early computer vision development. The way computers perceive things may not be the same as what we consider to be the human visual process. In chronological order, computer vision methods can be roughly divided into three stages: 1. rule models established based on expert-designed visual processes; 2. statistical models based on hand-crafted features designed by experts; 3. representation learning paradigm driven by data and computing power. In terms of performance, we observe an interesting phenomenon: the less prior knowledge of humans is utilized in the development of computer vision, whether it is rules or mathematics (or we consider rules as a simplified mathematical tool), the more computing power and data are used. But this is not a phenomenon unique to computer vision. We will find similar laws in other fields in the future.

\subsection{Natural Language Processing}

The origin of natural language processing can be traced back to the early 20th century, with figures such as Baudouin de Courtenay~\cite{Rothstein1972ABD} and Ferdinand de Saussure~\cite{saussure1916course} approaching language as a system of symbols with inherent rules, from a psychological and cognitive perspective. Over time, language research gradually departed from the human and social context, abstracting the study to rules of symbols, which had a profound impact on later natural language processing. Generally speaking, natural language processing has gone through three eras: rule-based, probabilistic statistical models, and deep learning.

The first era was rule-based language analysis, which lasted from the 1950s to the 1980s. The basic idea was to use expert-designed rule sets related to specific domains to enable computers to understand natural language, with the primary focus on syntax and semantics analysis. Noam Chomsky's book "Syntactic Structures"~\cite{Chomsky57a} was a representative work, which enabled computers to analyze sentences according to given rules, such as constructing a syntactic parse tree for a sentence. However, the complexity of the real world made the artificial rules an abstract and simplified process, detached from the practical application of language, and inadequate for understanding language in real-world situations. Similarly to the study of combination rules in computer vision, both fields depend entirely on experts' abstraction of the world, and their detachment from real situations makes it difficult to have practical value.

In the 1970s, statistical methods gradually became the main tool in natural language processing. Probability was used to represent words, rather than artificially defined language rules. The core idea was that the probability of a sentence equals the joint probability of each word that makes it up, based on the Markov assumption on sentences. Representative works include the general N-gram model~\cite{cavnar1994ngrambased} and the Hidden Markov Model (HMM)~\cite{HMM1991} used in speech recognition. However, statistical models invariably require prior assumptions, and then design models to fit the data, such as the typical example of topic models. These assumptions may not match the real-world situation, and the performance of the models is limited.

Since the advent of the deep learning era, natural language processing has made remarkable breakthroughs, thanks to big data and large-scale models, completely abandoning the rule-based and statistical model research paradigm. Early classic works in this stage include Word2vec, which achieved a breakthrough in word representation. From the perspective of model structure, the success of Seq2seq~\cite{seq2seq2014} made the model universality initially formed. With the emergence of Transformer~\cite{Attention2017}, universal models entered a new stage, directly leading to large-scale pre-training, becoming the unified mode of natural language processing. Recently, the appearance of ChatGPT has again significantly improved the performance of natural language processing. We notice that compared to the previous GPT-3~\cite{NEURIPS2020_1457c0d6}, the new version fundamentally differs by introducing large-scale human feedback mechanisms~\cite{lambert2022illustrating}, where human knowledge is directly fed back to the model to help it iterate.

Similar to the development of computer vision, natural language processing has undergone a development process from rules to data-driven approaches. However, the uniqueness of language deserves our attention, and we must recognize that language is one of the reasons why people become people. Language is not an independent tool or system detached from humans. Using strong rules or data statistics based on strong prior assumptions cannot explain human language, making it difficult to understand human language.

\subsection{Machine Learning}

In 1952, Arthur Samuel proposed the concept of machine learning, which is "a field of study that gives computers the ability to learn without being explicitly programmed." He demonstrated the possibility of machines outperforming humans by developing a checkers program. In pursuit of this goal, the field of machine learning has experienced ups and downs over the past seventy years, roughly divided into research directions based on neural networks and statistics.

In 1957, Rosenblatt proposed the perceptron~\cite{rosenblatt1958perceptron}, which can be considered as the earliest prototype of today's deep learning models and can be viewed as a biomimetic structure of a neuron. In 1969, Marvin Minsky raised the famous XOR problem, stating that the perceptron is ineffective on linearly inseparable data distributions. Researchers of neural networks entered a winter period until 1980, when Werbos proposed the multilayer perceptron (MLP). This model includes the core component of today's deep learning:  backpropagation (BP). At this point, the model level of deep learning has been initially completed. In 2006, Hinton and others' research on deep learning once again received attention. After nearly twenty years of rapid development, deep learning has completely changed research on AI in various fields. We need to pay attention to the fact that computing power and data are the most important variables in the current and past eras. The model itself is only a small part of the reason for success because even a simple MLP model can achieve high performance with enough computing power and data. Another direction of machine learning is based on statistics. In the two decades after the 1990s, statistical-based methods became the mainstream of machine learning. Representative works such as Boosting~\cite{Friedman2001}, Support Vector Machine (SVM)~\cite{cortes1995support}, Ensemble Learning~\cite{sagi2018ensemble}, Principal Component Analysis~\cite{Jolliffe1986} were born. Although they have a solid mathematical foundation, their performance cannot compete with today's deep learning.

Fundamentally speaking, current mathematical tools are not enough to model the real world. They are more of an abstraction and simplification of the real world. Typically, these mathematical tools are based on strong assumptions, and these empirical assumptions (guesses or judgments) cannot be rigorously proven in a scientific manner. Therefore, these seemingly rigorous mathematical tools are no different from expert rules in computer vision and natural language processing. Therefore, from the beginning of modeling, we must recognize the inadequacy of theory and rely on empirical judgments to make up for it. This is the underlying logic of the success of deep learning, which uses a data-driven approach to compensate for the inadequacy of mathematical tools and directly learns models of the real world from data.

%% file: tex/4_Phi.tex
\section{A Reflection from the Philosophy Perspective}

\subsection{The Necessity of Practice}

Although artificial intelligence has been developing for nearly a century, it still lacks a unified philosophical foundation. In particular, this discipline has broken down the opposition between natural sciences and social sciences since its inception, which means that it is itself a kind of unity. However, this unity is only formal at present.

The fundamental question about artificial intelligence is still the same: what is universal necessity? This universal necessity is actually the unity of the world, which is in motion and universally connected. In other words, it is the universality of contradictions. Truth or true knowledge is the unity of universality and particularity, and knowledge of universal necessity is obtained from limited experience. Sir Isaac Newton used mathematics to express this universality, which is the "Mathematical Principles of Natural Philosophy"~\cite{NewtonIsaac}. What is truth and what is universal necessity is the same question. This was originally a problem for metaphysics to solve. Its translation into metaphysics is not very accurate, and its original meaning is ontology or first philosophy. In other words, why are universal conclusions drawn from limited empirical materials reliable?

To solve this problem, philosophers have made remarkable efforts.  G.W.F. Hegel reconstructed metaphysics from the absolute, regarding truth as the alienation and return of absolute spirit. Specifically, Hegel's absolute is something that can only be grasped by pure thought, or it is pure thought itself. However, in reality, people always start from the limited to understand this infinite world. The so-called practice is endless, and understanding is endless, which negates Hegel's understanding of truth. Regarding artificial intelligence, Zhao Nanyuan proposed the theory of the generalized evolution of knowledge, which should demonstrate the universality and unity of knowledge. However, he said that the expansion of knowledge itself can prove its own effectiveness, which is equivalent to saying nothing. Moreover, the lack of unity in natural science also negates his conclusion.

Half a century ago, Mao Tse-tung's theory of practice and theory of contradictions solved the fundamental problem above from a philosophical level~\cite{Mao}. He attributed the answer to this problem to practice being the only criterion for testing truth at the practical level. The universality and immediacy of practice are the basis of the universality and unity of knowledge. His philosophical theory is based on the epistemology of Marxist dialectical materialism, which places practice in the first position, and believes that human cognition cannot be separated from practice. It rejects all erroneous theories that deny the importance of practice and make cognition separate from practice. We must emphasize the dependence of theory on practice, and the foundation of theory is practice, which in turn serves practice. The judgment of whether knowledge or theory is true is not based on subjective feelings, but on how the results of social practice are objectively determined. The standard of truth can only be social practice, and the viewpoint of practice is the first and fundamental viewpoint of dialectical materialism epistemology. Therefore, artificial intelligence is no exception and needs to be based on practice.

We can make a speculation here: the establishment of causality cannot be obtained by logical reasoning and mathematical statistics. The theory of practice provides the standard of truth, and its essence and causality are actually the same problem, both of which involve universality and necessity. The establishment of universality and necessity is ultimately determined by practice, with confirmation, refutation, and revision. Therefore, the establishment of causality cannot be obtained by logical reasoning and mathematical statistics. In addition, if mathematics and logic are used as the basis of causality, it is tantamount to saying that there is no world without mathematics and logic. Causality is the interaction of things, one thing causes another. Therefore, there are two prerequisites for the establishment of causality: 1. things exist; 2. there is a connection between things. Discovering and revealing this connection is the task of science. Mathematics and logic can be used to express causality, but they cannot become the main factor and say that causality exists because of mathematics and logic. The establishment of causality is ultimately determined by practice, and it should be noted that practice is infinite, and only the endless process of practice verification can prove its universality.

\subsection{Why will our artificial intelligence be doomed to fail?}


Why have our past AI attempts always failed? Because researchers have been eager to establish a knowledge system independent of humans, that is, to create a new intelligent species without human participation, allowing machines to make judgments independently of humans, whether based on rules or statistics and data-driven, fundamentally this is the vision. Specifically, AI simulates the thinking process of human beings from intuitive cognition to rational cognition, and mechanizes and mathematizes a part of human's actual thinking process. However, in the past, we focused on the conceptual deduction of this thinking process, such as Marr's visual process and Saussure's language symbols mentioned in Chapter 2, abstracting this thinking process. But this system did not incorporate human feedback into the development framework of AI technology. Instead, it treated humans as humans and machines as machines, as two independent systems (subject-object binary opposition), or as humans being equivalent to machines themselves (mechanical monism). In fact, machine activity is essentially mathematical reasoning without self-awareness, while practice is a conscious human activity. This subject-object separation system without human feedback also lacks practice, and machines cannot produce knowledge, let alone human-like intelligence. Therefore, human participation is an essential part of AI. ChatGPT is an accidental event in technological development, but it is a necessary result of historical development, because only by systematically incorporating human factors into the AI system can AI move towards inevitable success.

\subsection{The Necessity of Human-Machine Collaboration}

In the past, the philosophy of artificial intelligence was based on the subject-object dualism, where knowledge was the basis of knowledge, rather than practice. However, practice cannot be separated from feedback. Thus, implementing artificial intelligence requires human feedback, rather than simply allowing machines to independently achieve intelligence. Therefore, human-machine collaboration is inevitable, as it combines human cognitive abilities with machine capabilities, where human cognitive abilities are determined by human practice. Thus, human-machine collaboration is the necessity of artificial intelligence, rather than simply machine intelligence.

However, we must pay attention to the effectiveness of feedback. Taking recommendation systems as an example, as one of the most successful applications of artificial intelligence, the biggest problem is that users do not have true freedom of choice and decision-making. Although users can choose "dislike" or "not needed" in form, it is actually the imposition of the system's will on humans, which does not conform to the practice theory in reality.

\subsection{The Necessity of Computing Power}

We need to recognize the role of computers. Without computers, there would be no artificial intelligence. Both computing power and algorithms are essential for computation. In terms of computers, we have never demanded more than the computing advantage and cannot demand more. This is the reason why artificial intelligence development has faced problems, as we often demand more from computers than just computation, such as independent consciousness. The development of artificial intelligence technology must always be centered on computing power, even if there are more powerful computing tools in the future, such as quantum computers. We can analyze the fundamental contradictions of AI in two aspects: first, computing power is constant, so we can only use human knowledge to improve the computational performance of intelligent machines; second, human knowledge is too complex and difficult to teach machines through computation, so computing power can continue to expand. The result is that with the expansion of computing power, a certain level can be reached, such as when quantum computers mature, and the part of human knowledge that cannot currently be achieved through computation may become possible. Thus, the chain of human knowledge, algorithms, and computing power can spiral upwards. Therefore, in light of these real contradictions, development centered on computing power is an inevitable choice.

%% file: tex/5_social.tex
\section{Artificial Intelligence and Future Society}

The "practice" of artificial intelligence is to participate in the social production and life practice of human beings, that is, it is a part of human social production and life practice. Artificial intelligence acts as a tool and means in social production and life. From this perspective, it has no essential difference from the mechanical machines invented by humans before, which is its universality. Artificial intelligence is the product of deepening and upgrading of labor division, but unlike previous technological innovations, it has its own particularity. AI represents the replacement of mental labor for physical labor, creative labor for simple repetitive labor, and the shift from being controlled by the production process to controlling the entire production process, marking a new leap in human social productivity. This creates a more solid material basis for higher and more comprehensive human development and ultimately produces a more efficient and harmonious social form.

Artificial intelligence is rooted in human social production and life practice, and human social production and life practice are the fundamental driving force for promoting the research and development of artificial intelligence. Therefore, we can predict that population size and production scale will give practical advantages, and societies with this form will win in the evolution of artificial intelligence.

In the era of artificial intelligence technology, what is the role of human beings? First of all, we need to answer a question: what can humans do and what can't they do? The human brain can use conceptual thinking to process experience and generate new knowledge, but the data that the human brain can process is limited. Secondly, we need to answer another question: what can machines do and what can't they do? We all know that machines can perform efficient calculations, and anything that can be calculated can be done using a computer. However, machines cannot practice, so they cannot judge the truth of knowledge and may even forge knowledge. Therefore, future humans need to constantly provide new knowledge, ultimately serving three aspects: 1. increase computing power; 2. improve algorithms; 3. provide feedback to machines. Future social division of labor will also revolve around these three points. On one hand, it will eliminate old divisions of labor, and on the other hand, it will deepen new divisions of labor, especially in the development of intellectual labor. Artificial intelligence undoubtedly will expand the scope, breadth, and depth of social division of labor.

%% file: tex/conclusion.tex
\section{Conclusion}

This article summarizes the technological development and evolution of artificial intelligence over the past century from both technical and philosophical perspectives. We found that the development paths of different directions are generally similar, and the profound philosophical logic behind the failures and successes has never been valued in the field of AI, nor has it been systematically analyzed. It should be noted that AI is the combination of natural science and social science, not just machine intelligence, but an extension of human intelligence. Therefore, it is necessary to understand AI from different perspectives in order to avoid blind development. We call for attention to the establishment of the philosophical foundation of AI, and the urgent need for a unified and time-tested philosophical system in the AI field to ultimately guide the development of technology.

%% file: main.bbl
\newcommand{\etalchar}[1]{$^{#1}$}
\begin{thebibliography}{LCvWH22}

\bibitem[BMR{\etalchar{+}}20]{NEURIPS2020_1457c0d6}
Tom Brown, Benjamin Mann, Nick Ryder, Melanie Subbiah, Jared~D Kaplan, Prafulla
  Dhariwal, Arvind Neelakantan, Pranav Shyam, Girish Sastry, Amanda Askell,
  Sandhini Agarwal, Ariel Herbert-Voss, Gretchen Krueger, Tom Henighan, Rewon
  Child, Aditya Ramesh, Daniel Ziegler, Jeffrey Wu, Clemens Winter, Chris
  Hesse, Mark Chen, Eric Sigler, Mateusz Litwin, Scott Gray, Benjamin Chess,
  Jack Clark, Christopher Berner, Sam McCandlish, Alec Radford, Ilya Sutskever,
  and Dario Amodei.
\newblock Language models are few-shot learners.
\newblock In H.~Larochelle, M.~Ranzato, R.~Hadsell, M.F. Balcan, and H.~Lin,
  editors, {\em Advances in Neural Information Processing Systems}, volume~33,
  pages 1877--1901. Curran Associates, Inc., 2020.

\bibitem[Cho57]{Chomsky57a}
Noam Chomsky.
\newblock {\em Syntactic Structures}.
\newblock Mouton and Co., The Hague, 1957.

\bibitem[CT94]{cavnar1994ngrambased}
William~B. Cavnar and John~M. Trenkle.
\newblock {N}-gram-based text categorization.
\newblock In {\em Proceedings of {SDAIR}-94, 3rd Annual Symposium on Document
  Analysis and Information Retrieval}, pages 161--175, Las Vegas, US, 1994.

\bibitem[CV95]{cortes1995support}
C.~Cortes and V.~Vapnik.
\newblock Support vector networks.
\newblock {\em Machine Learning}, 20:273--297, 1995.

\bibitem[DBK{\etalchar{+}}20]{ViT2020}
Alexey Dosovitskiy, Lucas Beyer, Alexander Kolesnikov, Dirk Weissenborn,
  Xiaohua Zhai, Thomas Unterthiner, Mostafa Dehghani, Matthias Minderer, Georg
  Heigold, Sylvain Gelly, Jakob Uszkoreit, and Neil Houlsby.
\newblock An image is worth 16x16 words: Transformers for image recognition at
  scale, 2020.

\bibitem[dS83]{saussure1916course}
Ferdinand de~Saussure.
\newblock {\em Course in General Linguistics}.
\newblock Duckworth, London, [1916] 1983.
\newblock (trans. Roy Harris).

\bibitem[DT05]{Triggs_hog}
N.~Dalal and B.~Triggs.
\newblock Histograms of oriented gradients for human detection.
\newblock In {\em 2005 IEEE Computer Society Conference on Computer Vision and
  Pattern Recognition (CVPR'05)}, volume~1, pages 886--893 vol. 1, 2005.

\bibitem[FGMR09]{felzenszwalb2009object}
Pedro~F Felzenszwalb, Ross~B Girshick, David McAllester, and Deva Ramanan.
\newblock Object detection with discriminatively trained part-based models.
\newblock {\em IEEE transactions on pattern analysis and machine intelligence},
  32(9):1627--1645, 2009.

\bibitem[Fri01]{Friedman2001}
Jerome~H. Friedman.
\newblock {Greedy function approximation: A gradient boosting machine.}
\newblock {\em The Annals of Statistics}, 29(5):1189 -- 1232, 2001.

\bibitem[HOT06]{Hinton2006}
Geoffrey Hinton, Simon Osindero, and Yee-Whye Teh.
\newblock A fast learning algorithm for deep belief nets.
\newblock {\em Neural computation}, 18:1527--54, 08 2006.

\bibitem[Jol86]{Jolliffe1986}
I.T. Jolliffe.
\newblock {\em Principal Component Analysis}.
\newblock Springer Verlag, 1986.

\bibitem[JR91]{HMM1991}
B.~H. Juang and L.~R. Rabiner.
\newblock Hidden markov models for speech recognition.
\newblock {\em Technometrics}, 33(3):251--272, 1991.

\bibitem[KSH12]{Alex2012}
Alex Krizhevsky, Ilya Sutskever, and Geoffrey~E Hinton.
\newblock Imagenet classification with deep convolutional neural networks.
\newblock In F.~Pereira, C.J. Burges, L.~Bottou, and K.Q. Weinberger, editors,
  {\em Advances in Neural Information Processing Systems}, volume~25. Curran
  Associates, Inc., 2012.

\bibitem[LCvWH22]{lambert2022illustrating}
Nathan Lambert, Louis Castricato, Leandro von Werra, and Alex Havrilla.
\newblock Illustrating reinforcement learning from human feedback (rlhf).
\newblock {\em Hugging Face Blog}, 2022.
\newblock https://huggingface.co/blog/rlhf.

\bibitem[LSP06]{Lazebnik06}
S.~Lazebnik, C.~Schmid, and J.~Ponce.
\newblock Beyond bags of features: Spatial pyramid matching for recognizing
  natural scene categories.
\newblock In {\em 2006 IEEE Computer Society Conference on Computer Vision and
  Pattern Recognition (CVPR'06)}, volume~2, pages 2169--2178, 2006.

\bibitem[Mar82]{Marr1982}
David Marr.
\newblock {\em Vision: A Computational Investigation into the Human
  Representation and Processing of Visual Information}.
\newblock Henry Holt and Co., Inc., New York, NY, USA, 1982.

\bibitem[MMRS06]{McCarthyMRS06}
John McCarthy, Marvin Minsky, Nathaniel Rochester, and Claude~E. Shannon.
\newblock A proposal for the dartmouth summer research project on artificial
  intelligence, august 31, 1955.
\newblock {\em AI Magazine}, 27(4):12--14, 2006.

\bibitem[NAJ29]{NewtonIsaac}
Isaac Newton, Motte Andrew, and Machin John.
\newblock {\em The Mathematical Principles of Natural Philosophy}.
\newblock Springer Verlag, 1729.

\bibitem[RdCS72]{Rothstein1972ABD}
Robert~A. Rothstein, Jan~Baudouin de~Courtenay, and Edward~B. Stankiewicz.
\newblock A baudouin de courtenay anthology : the beginnings of structural
  linguistics.
\newblock {\em Slavic and East European Journal}, 16:495, 1972.

\bibitem[Ros58]{rosenblatt1958perceptron}
Frank Rosenblatt.
\newblock The perceptron: a probabilistic model for information storage and
  organization in the brain.
\newblock {\em Psychological review}, 65(6):386, 1958.

\bibitem[SB92]{Binford1992}
Hiroaki Sato and Thomas~O. Binford.
\newblock On finding the ends of straight homogeneous generalized cylinders.
\newblock In {\em {IEEE} Computer Society Conference on Computer Vision and
  Pattern Recognition, {CVPR} 1992, Proceedings, 15-18 June, 1992, Champaign,
  Illinois, {USA}}, pages 695--698. {IEEE}, 1992.

\bibitem[Sha20]{CV50year}
Linda~G. Shapiro.
\newblock Computer vision: the last 50 years.
\newblock {\em International Journal of Parallel, Emergent and Distributed
  Systems}, 35(2):112--117, 2020.

\bibitem[SR18]{sagi2018ensemble}
Omer Sagi and Lior Rokach.
\newblock Ensemble learning: A survey.
\newblock {\em Wiley Interdisciplinary Reviews: Data Mining and Knowledge
  Discovery}, 8(4):e1249, 2018.

\bibitem[SVL14]{seq2seq2014}
Ilya Sutskever, Oriol Vinyals, and Quoc~V. Le.
\newblock Sequence to sequence learning with neural networks, 2014.

\bibitem[Tt51]{Mao}
Mao Tse-tung.
\newblock {\em Selected Works of Mao Tse-tung}.
\newblock Springer Verlag, 1951.

\bibitem[VJ01]{ViolaJones}
P.~Viola and M.~Jones.
\newblock Rapid object detection using a boosted cascade of simple features.
\newblock In {\em Proceedings of the 2001 IEEE Computer Society Conference on
  Computer Vision and Pattern Recognition. CVPR 2001}, volume~1, pages I--I,
  2001.

\bibitem[VSP{\etalchar{+}}17]{Attention2017}
Ashish Vaswani, Noam Shazeer, Niki Parmar, Jakob Uszkoreit, Llion Jones,
  Aidan~N. Gomez, Lukasz Kaiser, and Illia Polosukhin.
\newblock Attention is all you need, 2017.

\bibitem[ZM06]{Zhu2006}
Song-Chun Zhu and David Mumford.
\newblock A stochastic grammar of images.
\newblock {\em Found. Trends. Comput. Graph. Vis.}, 2(4):259–362, jan 2006.

\end{thebibliography}
